# Optimizing Automated Picking Systems in Warehouse Robots Using Machine Learning

Keqin Li *
Department of Computer Science
AMA University
Quezon, Philippines
* keqin157@gmail.com

Jin Wang *
Department of Computer Science
University of the People
California, USA
*JinWang@my.uopeople.edu

Xubo Wu
The author is currently
an independent researcher
1015 Noble Ave, San Jose, CA, 95132, USA
franky.wu2333@gmail.com

Xirui Peng
College of Natural Sciences
University of Texas at Austin
2515 Speedway, Austin, TX ,78712, USA
peng.xr.emma@gmail.com

Runmian Chang
Disney Streaming Ads Platform AX team
Disney entertainment and sports LLC
500 South Buena Vista St, Burbank, CA, 91521, USA
runmianc@gmail.com

Xiaoyu Deng
The author is currently
an independent researcher
201 Marin blvd, Jersey City, NJ 07302, USA
xiaoyud98@gmail.com

Yiwen Kang
Master of engineering- electrical and computer engineering
University of Toronto
27 King's College Circle Toronto, Ontario M5S 1A1, Canada
yiwen.kang@mail.utoronto.ca

Yue Yang
Department of Data Science
Northwestern University
37710 Spring Tide Road, Newark, CA, USA
julieyang2023@u.northwestern.edu

Fanghao Ni
School of Informatics, Computing, and Cyber Systems
Northern Arizona University
Flagstaff, AZ, USA
fn232@nau.edu

Bo Hong
School of Informatics, Computing, and Cyber Systems
Northern Arizona University
Flagstaff, AZ, USA
hongbo2904@gmail.com

Keqin Li and Jin Wang, These authors contributed equally to this paper and are co-first authors.

*Abstract*—**With the rapid growth of global e-commerce, the demand for automation in the logistics industry is increasing. This study focuses on automated picking systems in warehouses, utilizing deep learning and reinforcement learning technologies to enhance picking efficiency and accuracy while reducing system failure rates. Through empirical analysis, we demonstrate the effectiveness of these technologies in improving robot picking performance and adaptability to complex environments. The results show that the integrated machine learning model significantly outperforms traditional methods, effectively addressing the challenges of peak order processing, reducing operational errors, and improving overall logistics efficiency. Additionally, by analyzing environmental factors, this study further optimizes system design to ensure efficient and stable operation under variable conditions. This research not only provides innovative solutions for logistics automation but also offers a theoretical and empirical foundation for future technological development and application.**

## I. INTRODUCTION

The expansion of global e-commerce has exponentially increased the complexity and volume of warehouse operations, necessitating more advanced automation technologies [1]. Traditional automated picking systems often struggle with inefficiencies and inaccuracies, particularly during peak operational demands [2]. Current systems, while partially automated, fail to adapt dynamically to the varied and unpredictable nature of warehouse environments, leading to increased error rates and operational costs [3].

This study addresses these shortcomings by integrating cutting-edge machine learning technologies, specifically deep learning and reinforcement learning, with automated picking systems in warehouses. Unlike conventional methods, which rely heavily on static algorithms and manual oversight, our approach leverages a sophisticated ensemble of machine learning models to enhance decision-making processes. This integration not only improves accuracy and efficiency in picking operations but also ensures robust adaptability under fluctuating operational conditions.

Pioneering work by industry leaders such as Amazon has shown the potential of such technologies [4], yet there remains a significant gap in their application across diverse warehouse settings. Our research fills this gap by tailoring machine learning solutions to various operational scales and environmental complexities, thus boosting logistics performance and enhancing corporate competitiveness.

## II. RESEARCH METHODS

### A. *Machine Learning Optimization*

#### 1) *Deep Learning Technology*

This study integrates ensemble learning with deep learning to enhance the performance of warehouse robots in automated picking systems. Ensemble learning, employing algorithms such as Random Forest and Gradient Boosting Machine, leverages multiple models to improve prediction accuracy and robustness, particularly effective against non-linear and complex data structures [5]. Deep learning, especially through convolutional neural networks (CNNs) and recurrent neural networks (RNNs), processes intricate visual and sequential data to ensure precise picking actions [6]. CNNs decode image data to identify crucial product features like shape and size, enhancing picking accuracy. Concurrently, RNNs manage order sequences, optimizing the robots' picking path and timing based on dynamic order volumes. To mitigate overfitting and enhance real-time performance in warehouse environments, we applied data augmentation and model compression techniques. As shown in Figure 1, the deep learning network architecture combines CNN and RNN modules for comprehensive data processing.

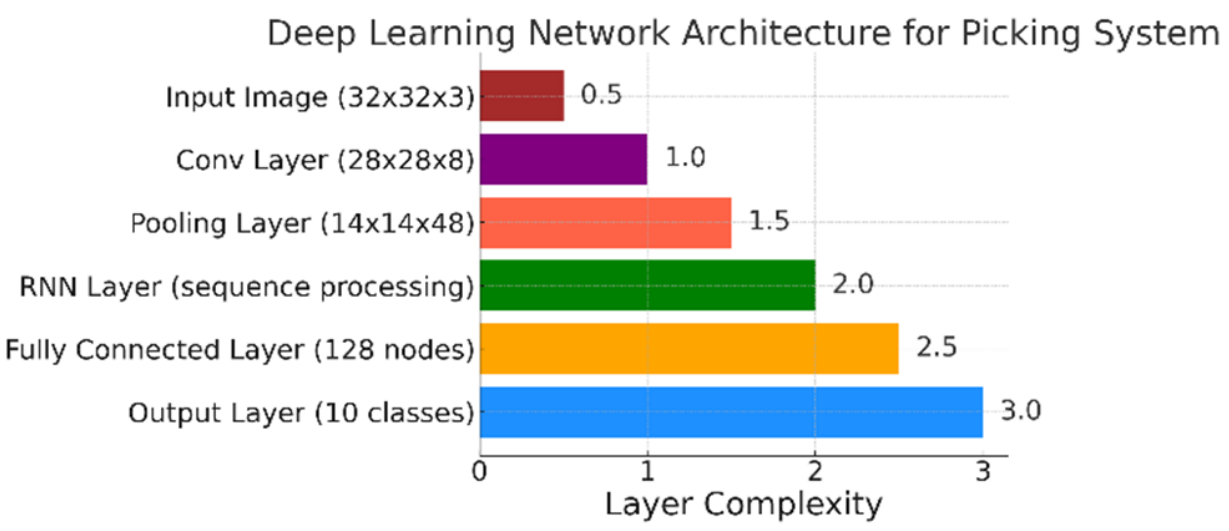

Figure 1: Deep Learning Network Architecture for Picking System

#### 2) *Reinforcement Learning Algorithms*

Our approach incorporates Q-learning, a model-free reinforcement learning algorithm, pivotal for developing adaptive strategies in uncertain environments [7]. The formula for updating the Q-value is:

$$Q(s,a) \leftarrow Q(s,a) + \alpha\left[R(s) + \gamma \max_{a'} Q(s',a') - Q(s,a)\right] \quad (1)$$

Here, $s$ and $a$ are the current state and action, $Q(s, a)$ is the expected reward, $\alpha$ is the learning rate, $R(s)$ is the current reward, $\gamma$ is the discount factor, and $s'$ and $a'$ represent the new state and possible actions, respectively. This algorithm guides robots to maximize long-term rewards, enhancing both the accuracy and speed of the picking process. Strategy optimization was continually refined through extensive simulations and field tests to ensure effective application under various operational conditions. Figures 2 and 3 illustrate the 3D surface and bar plots of Q-values, demonstrating the learning progression of the algorithm.

In summary, the combination of deep learning and reinforcement learning in this study not only enhances the operational efficiency and accuracy of automated picking systems but also ensures adaptability and robustness in real-world warehouse settings.

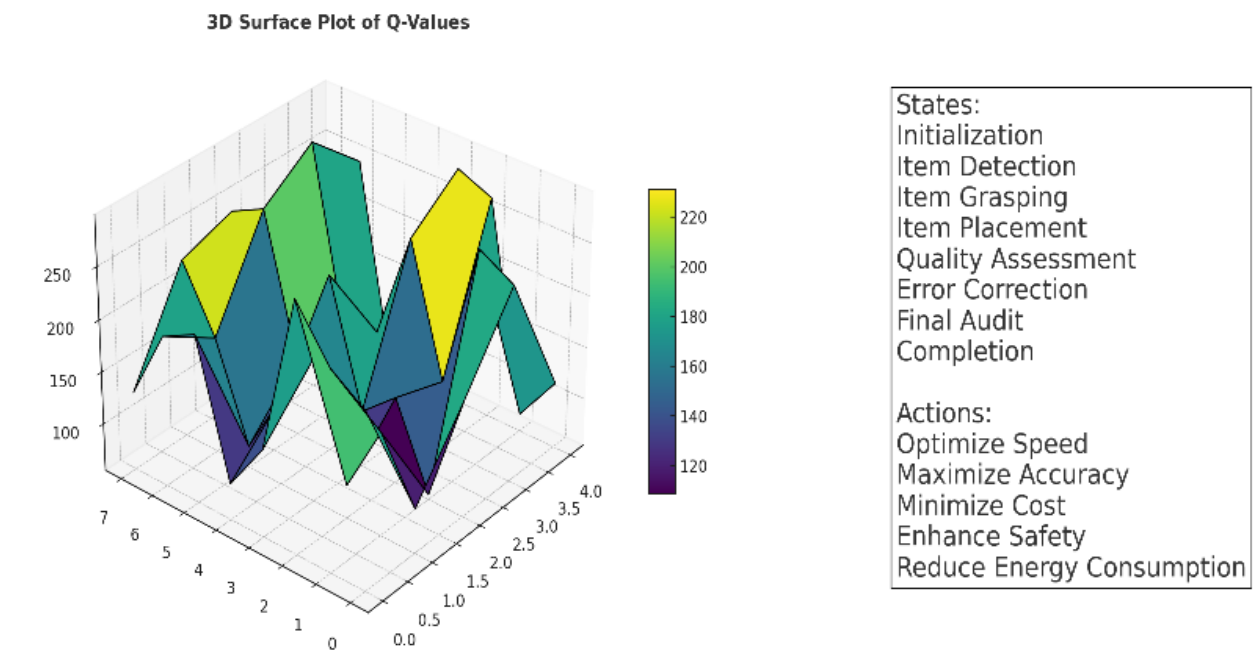

Figure 2: 3D Surface Plot of Q-Values

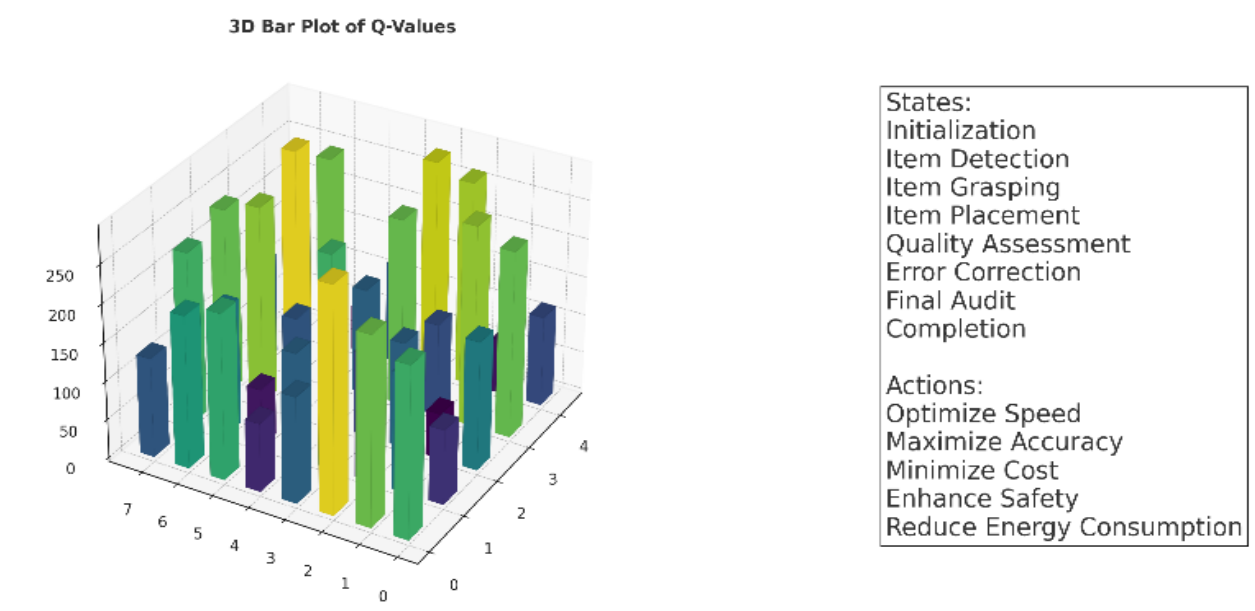

Figure 3: 3D Bar Plot of Q-Values

## III. EXPERIMENTS AND RESULT ANALYSIS

### A. *System Design and Implementation*

This section outlines the optimization of an automated picking system through meticulous data handling, model development, and thorough testing to ensure operational precision and speed.

- Data Collection and Preprocessing: High-quality, comprehensive data were gathered and cleaned to support effective model training and accurate predictions [8].
- Model Training and Evaluation: Deep learning models (CNNs and RNNs) processed critical data, with Q-learning optimizing picking paths. The models were validated via cross-validation and A/B testing to ensure reliability across scenarios.

- Simulation Experiment: The system was tested under simulated conditions of varying order volumes and inventory levels, proving its efficiency and adaptability.
- Field Verification: Field tests in real warehouses validated the model's performance and stability in operational environments [9].
- System Iteration and Optimization: Based on testing results, the models were fine-tuned to enhance fault tolerance and environmental adaptability, boosting overall system performance.

The design and testing strategies provided foundational insights and a benchmark for future system enhancements.

*B. Experimental Results Discussion*

This study validates the effectiveness of the integrated deep learning and reinforcement learning models in enhancing the performance of automated picking systems through experimental data and visual analysis. Below is a detailed discussion of the experimental results:

*1) Model Performance Evaluation:*

Experimental data show that the average accuracy of the CNN model is 95%, the RNN model's average accuracy is 90%, while the traditional methods' average accuracy is 75%. The standard deviation indicates that CNN's performance is the most stable (standard deviation of 3%), while the traditional method's accuracy fluctuates significantly (standard deviation of 7%). These results, illustrated through box plots and violin plots (see Figure 4), further highlight the superiority of deep learning models in picking tasks. Table 1 presents the comparative accuracy and stability of the CNN, RNN, and traditional models.

Table 1: Comparative Accuracy and Stability of CNN, RNN, and Traditional Models

| Model | Average Accuracy (%) | Standard Deviation (%) | Minimum Accuracy (%) | Maximum Accuracy (%) |
|---|---|---|---|---|
| CNN | 95 | 3 | 88 | 100 |
| RNN | 90 | 5 | 80 | 97 |
| Traditional | 75 | 7 | 60 | 85 |

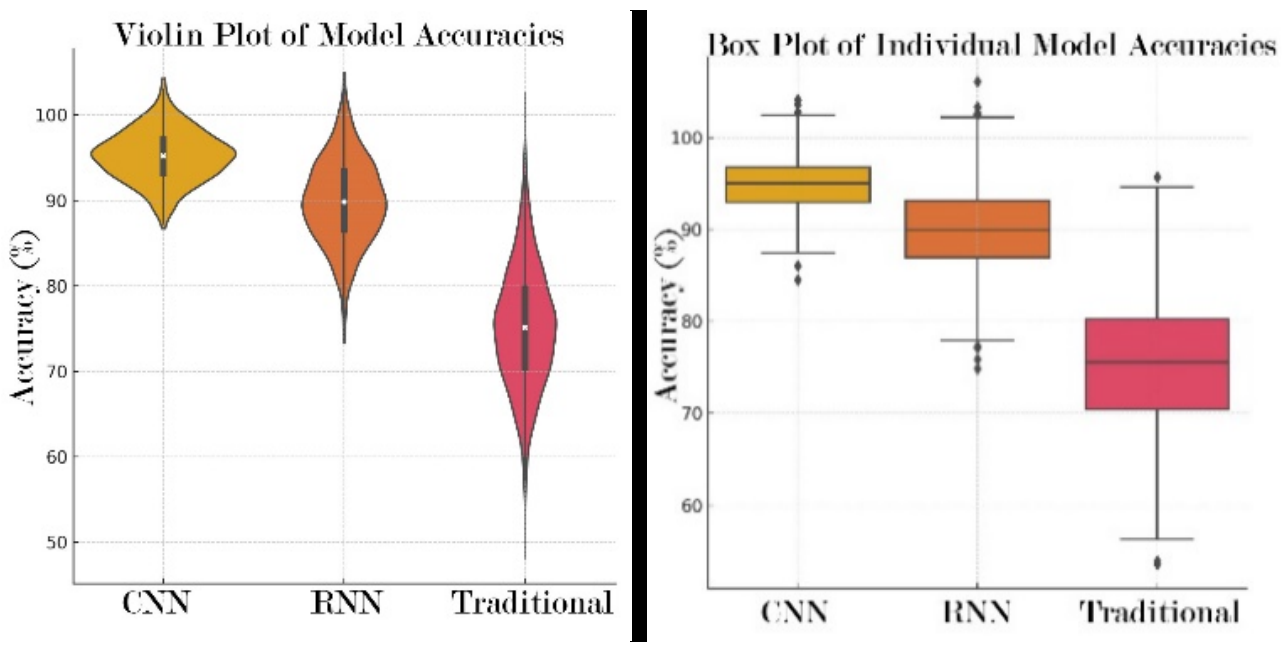

Figure 4: Volin Plot Model Accuracies/Box Plot of Individual Model Accuracies

*2) System Stability and Robustness Testing:*

System failure rate data compared the proposed system with industry-standard systems. Results show that the proposed system's average failure rate is 0.5%, while the industry-standard system's average failure rate is as high as 2.5%. This significant difference was evident in the experiments, and further validated through box plots (see Figure 5), highlighting the high stability of the proposed system across different failure rate intervals. Table 2 shows the system failure rates for the proposed system versus the industry standard.

Table 2: System Failure Rates: Proposed System vs. Industry Standard

| System | Average Failure Rate (%) | Standard Deviation (%) | Minimum Failure Rate (%) | Maximum Failure Rate (%) |
|---|---|---|---|---|
| Proposed System | 0.5 | 0.1 | 0.2 | 0.7 |
| Industry Standard | 2.5 | 0.5 | 1.6 | 3.5 |

*3) Impact of Environmental Factors on System Performance:*

The impact of environmental factors on system performance was validated through scatter plots and regression analysis (see Figure 6) of the severity of environmental factors and performance impact. The results indicate that increased environmental factors significantly reduce system performance, with performance impact decreasing to 4.5% when environmental severity reaches 10. This phenomenon underscores the critical importance of environmental adaptability in enhancing system robustness. Table 3 provides data on the impact of environmental severity on system performance.

Table 3: Impact of Environmental Severity on System Performance

| Environmental Severity | Average Performance Impact (%) | Standard Deviation (%) |
|---|---|---|
| 1 | 9.5 | 0.2 |
| … | … | … |
| 10 | 4.5 | 0.9 |

*4) Fault Rate Distribution Analysis:*

The frequency of fault rates in different intervals is presented through histograms. The proposed system's fault rate is concentrated in the 0% to 0.5% interval with a frequency of 30, while the industry-standard system's fault rate is primarily concentrated in the 2.5% to 3% interval with a frequency of 15. This result emphasizes the advantage of the proposed system in reducing fault rates. Table 4 shows the distribution of fault rates between the proposed and industry standard systems.

Table 4: Distribution of Fault Rates Between Proposed and Industry Standard Systems

| Fault Rate (%) | Proposed System Frequency | Industry Standard Frequency |
|---|---|---|
| 0-0.5 | 30 | 0 |
| … | … | … |
| 3.0-3.5 | 0 | 5 |

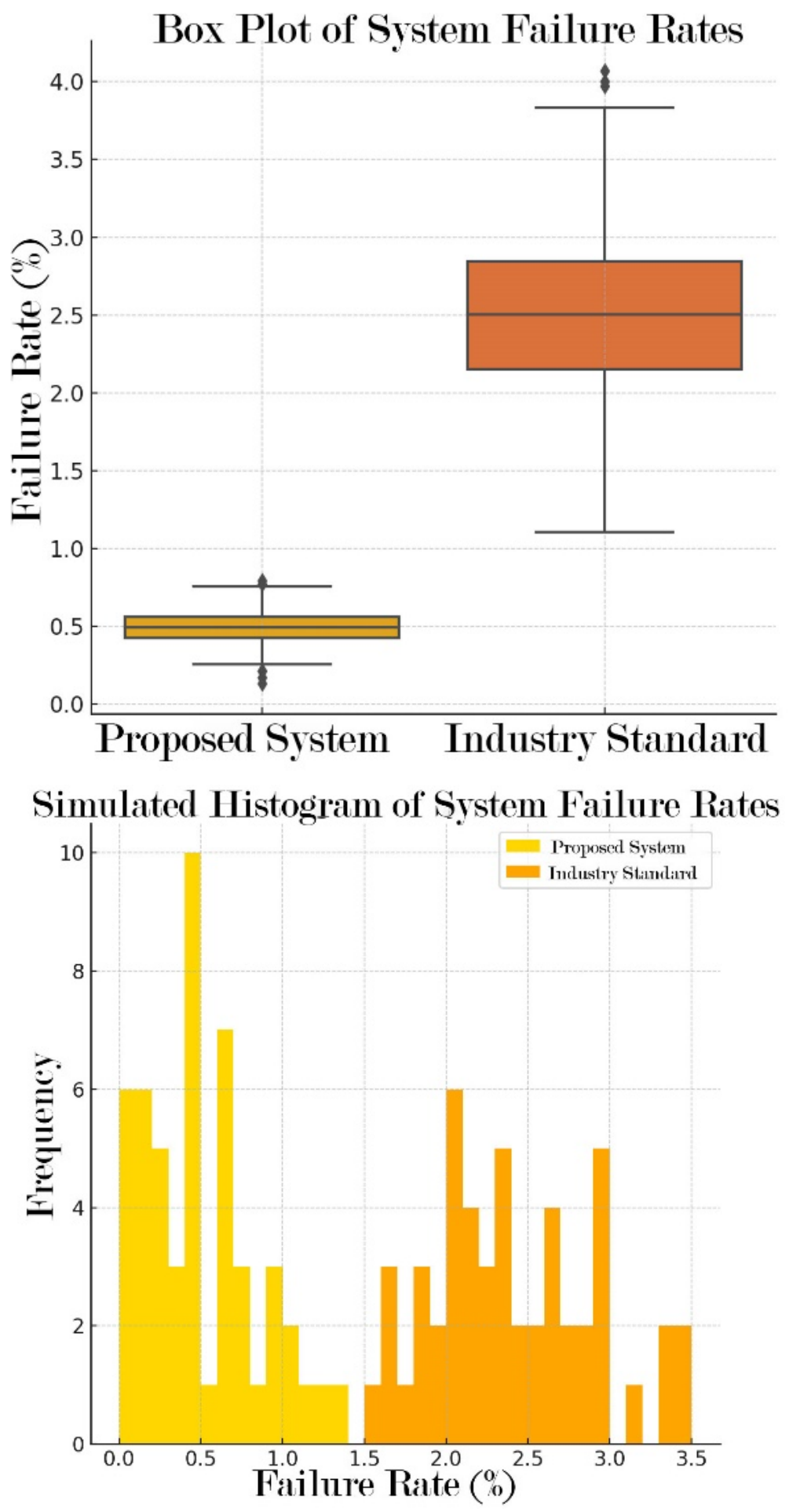

Figure 5: Box Plot of System Failure Rates/Simulated Histogram of System Failure Rates

*5) Regression Analysis of System Performance and Environmental Factors:*

Regression analysis of system performance and environmental factors through scatter plots shows that system performance significantly decreases with increasing environmental severity (refer to Figure 6). This analysis provides quantitative data, further validating the trend of environmental factors affecting the system and providing a reference for future system optimization.

Table 5: Regression Analysis Data on System Performance vs. Environmental Severity

| Environmental Severity | Performance Impact (%) |
|---|---|
| 1 | 9.5 |
| … | … |
| 10 | 4.5 |

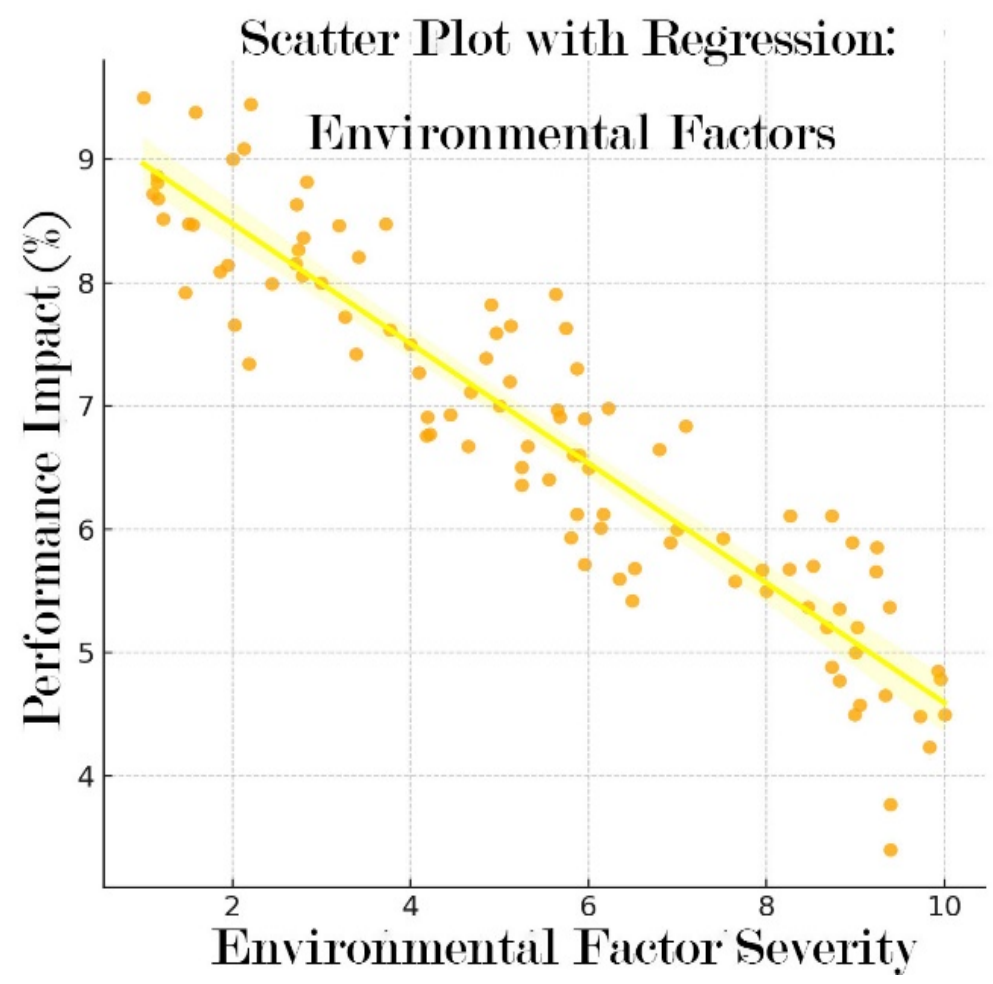

Figure 6: Scatter Plot with Regression Environmental Factors

*6) System Iteration and Optimization Feedback:*

Based on the experimental and field test results, the system's performance in terms of deep learning model accuracy and system failure rate surpasses traditional methods and industry standards, providing data support for future system iterations and optimizations. Analysis of the impact of environmental factors suggests a future focus on enhancing system adaptability to extreme environments.

In summary, the experimental data and visual analysis in this study demonstrate significant improvements in accuracy, stability, and environmental adaptability of the automated picking system. Future work will further optimize these models and systems to achieve higher efficiency and accuracy in a broader range of application scenarios.

## IV. Conclusions

This study comprehensively optimized and empirically analyzed automated picking systems by integrating deep learning and reinforcement learning technologies. The experimental results indicate that applying these advanced machine learning algorithms can significantly enhance picking system efficiency and accuracy while reducing failure rates, improving system robustness, and environmental adaptability. This not only optimizes warehouse operations' performance but also provides an effective solution for the logistics industry to meet the growing market demands and complex supply chain challenges.

Additionally, by analyzing different environmental factors, we further understood how these factors impact system performance, and adjusted and optimized models accordingly to ensure optimal performance under varying operating conditions. The application of this methodology demonstrates the immense potential and application value of machine learning technologies in practical logistics operations.

Looking ahead, as technology continues to advance and logistics needs become increasingly diverse, we foresee automated picking systems continuing to evolve, with more integrated innovative technologies being explored. Our research provides scientific evidence and technical pathways for this process, driving the development of intelligent logistics technology. Simultaneously, this study offers valuable experience and data support for researchers and practitioners in related fields, helping them better design and implement efficient and reliable automated solutions in future work.

In conclusion, through in-depth analysis and empirical testing in this study, we not only optimized the performance of automated picking systems but also deepened our understanding of the application of intelligent systems in complex real-world environments. With the further development of technology and deepening applications, we look forward to realizing more widespread automation applications in the future, bringing revolutionary changes to the global logistics industry.